%% file: main.tex
\documentclass[sigconf]{acmart}
\usepackage{soul}
\usepackage{graphicx}
\usepackage{subcaption}
\usepackage{color, colortbl}

\definecolor{Gray}{gray}{0.9}
\AtBeginDocument{%
  \providecommand\BibTeX{{%
    \normalfont B\kern-0.5em{\scshape i\kern-0.25em b}\kern-0.8em\TeX}}}

\setcopyright{acmlicensed}
\copyrightyear{2018}
\acmYear{2018}
\acmDOI{XXXXXXX.XXXXXXX}

\acmConference[KDD '24]{}{August,
  2024}{Barcelona, SPA}
\acmISBN{978-1-4503-XXXX-X/18/06}

\begin{document}

\title{A Bag of Tricks for Scaling CPU-based Deep FFMs to \\more than 300m Predictions per Second}
%\title{Scaling CPU-based Quantized Deep FFMs to 300m+ PPS}


\author{Bla\v{z} \v{S}krlj}
\affiliation{%
  \institution{Outbrain}
  \country{}}
\email{bskrlj@outbrain.com}

\author{Benjamin Ben-Shalom}
\affiliation{%
  \institution{Outbrain}
  \country{}}
\email{bbshalom@outbrain.com}

\author{Grega Ga\v{s}per\v{s}i\v{c}}
\affiliation{%
  \institution{Outbrain}
  \country{}}
\email{ggaspersic@outbrain.com}

\author{Adi Schwartz}
\affiliation{%
  \institution{Outbrain}
  \country{}}
\email{aschwartz@outbrain.com}

\author{Ramzi Hoseisi}
\affiliation{%
  \institution{Outbrain}
  \country{}}
\email{rhoseisi@outbrain.com}

\author{Naama Ziporin}
\affiliation{%
  \institution{Outbrain}
  \country{}}
\email{nziporin@outbrain.com}

\author{Davorin Kopi\v{c}}
\affiliation{%
  \institution{Outbrain}
  \country{}}
\email{dkopic@outbrain.com}

\author{Andra\v{z} Tori}
\affiliation{%
  \institution{Outbrain}
  \country{}}
\email{atori@outbrain.com}
\renewcommand{\shortauthors}{\v{S}krlj, et al.}

\begin{abstract}
Field-aware Factorization Machines (FFMs) have emerged as a powerful model for click-through rate prediction, particularly excelling in capturing complex feature interactions. In this work, we present an in-depth analysis of our in-house, Rust-based Deep FFM implementation, and detail its deployment on a CPU-only, multi-data-center scale. We overview key optimizations devised for both training and inference, demonstrated by previously unpublished benchmark results in efficient model search and online training.
Further, we detail an in-house weight quantization that resulted in more than an order of magnitude reduction in bandwidth footprint related to weight transfers across data-centres. We disclose the engine and associated techniques under an open-source license to contribute to the broader machine learning community. This paper showcases one of the first successful CPU-only deployments of Deep FFMs at such scale, marking a significant stride in practical, low-footprint click-through rate prediction methodologies.
\end{abstract}






\maketitle
\begin{figure}[t!]
    \centering
    \includegraphics[width=.97\linewidth]{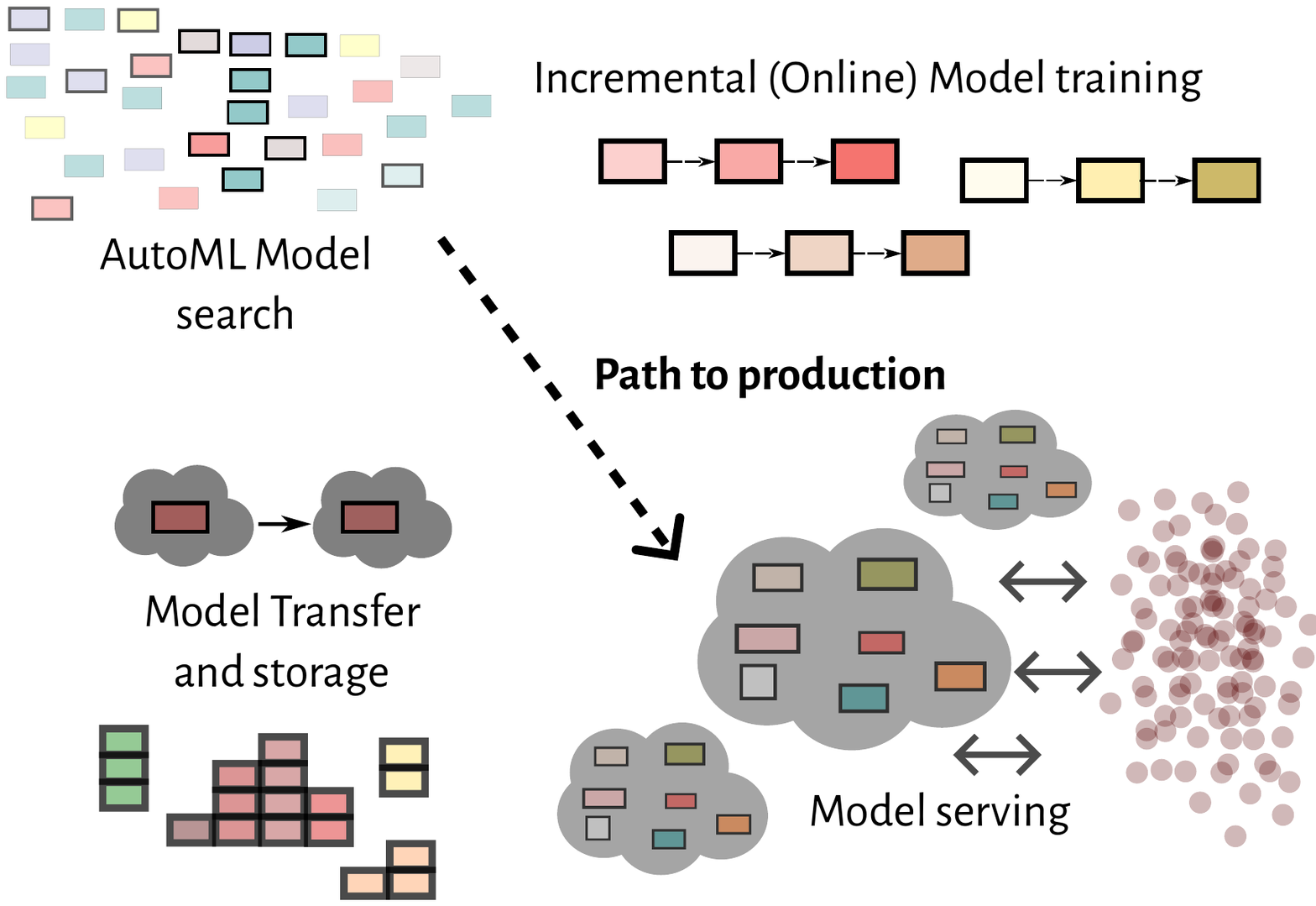}
    \caption{Overview of the key topics discussed in this paper. Performance optimizations that span model search (AutoML), online model training, storage, transfer and serving are discussed.}
    \label{fig:overview}
    \vspace{-0.3cm}
\end{figure}
\section{Introduction}
Design and development of machine learning approaches for the domain of \emph{recommendation systems} revolves around the interplay between scalability and approximation capability of classification and regression algorithms. Currently, many deployed recommendation engines rely on factorization machine-based approaches; this is mostly due to good trade-offs when it comes to scalability, maintainability and data scientists' involvement in building such models. Even though contemporary recommenders started to increasingly rely on language model-based techniques~\cite{zhang2023chatgpt}, utilizing factorization machines remains \emph{de facto} solution for large-scale "screening" of candidates that are to be served. Such candidates can include from unseen items (online stores), to movie recommendations, to ads~\cite{zhang2021deep,deldjoo2020recommender}. Scalability of factorization machines enables creation of real-time systems that handle hundreds of millions of requests in predictable and maintainable manner. In recent years, two main branches of methods have emerged. Approaches based on frameworks such as TensorFlow~\cite{tensorflow2015-whitepaper} and PyTorch~\cite{torch} enabled construction of highly expressive architectures that often require specialized hardware for efficient productization~\cite{song2019autoint,lian2018xdeepfm,cheng2016wide,guo2017deepfm}. CPU-only, single instance -- single pass alternatives are fewer, and revolve around highly optimized C++ or Rust-based approaches that exploit consumer hardware as much as possible. The latter is the main focus of this paper (overview in Figure~\ref{fig:overview}). 

\section{Fwumious Wabbit (FW) - an overview}
We proceed with a discussion of Fwumious Wabbit (FW), an in-house, Rust-based factorization machine-based system currently used in production for large-scale recommendation\footnote{The engine with main implementations discussed in this paper is freely available as \url{https://github.com/outbrain/fwumious_wabbit}.}.

\subsection{Origins of FW and Vowpal Wabbit (VW)}
The FW derives from Vowpal Wabbit (VW) \cite{bietti2018a}, a high-performance, scalable open-source ML system recognized for its efficiency on large datasets \footnote{\url{https://vowpalwabbit.org/}}. While VW primarily uses logistic regression for tasks like click-through rate prediction, it lacks readily available advanced extensions found in the domain of factorization machines.
One of the more expressive variations of factorization machines are the Field-aware Factorization Machines (FFMs), described in detail in the works of Juan et al.~\cite{juan2017field,juan2016field}. Building on this foundation, we enhanced the FFM architecture by integrating elements of deep learning. Specifically, a multi-layer perceptron (MLP)-like structure in conjunction with the traditional FFM (and logistic regression) components. The architecture's computational complexity, a notable challenge, contributes to its rarity in existing benchmarks. When implemented in standard frameworks like TensorFlow, the architecture struggles to scale effectively for practical use.

\begin{figure}[t!]
    \centering
    \includegraphics[width=.95\linewidth]{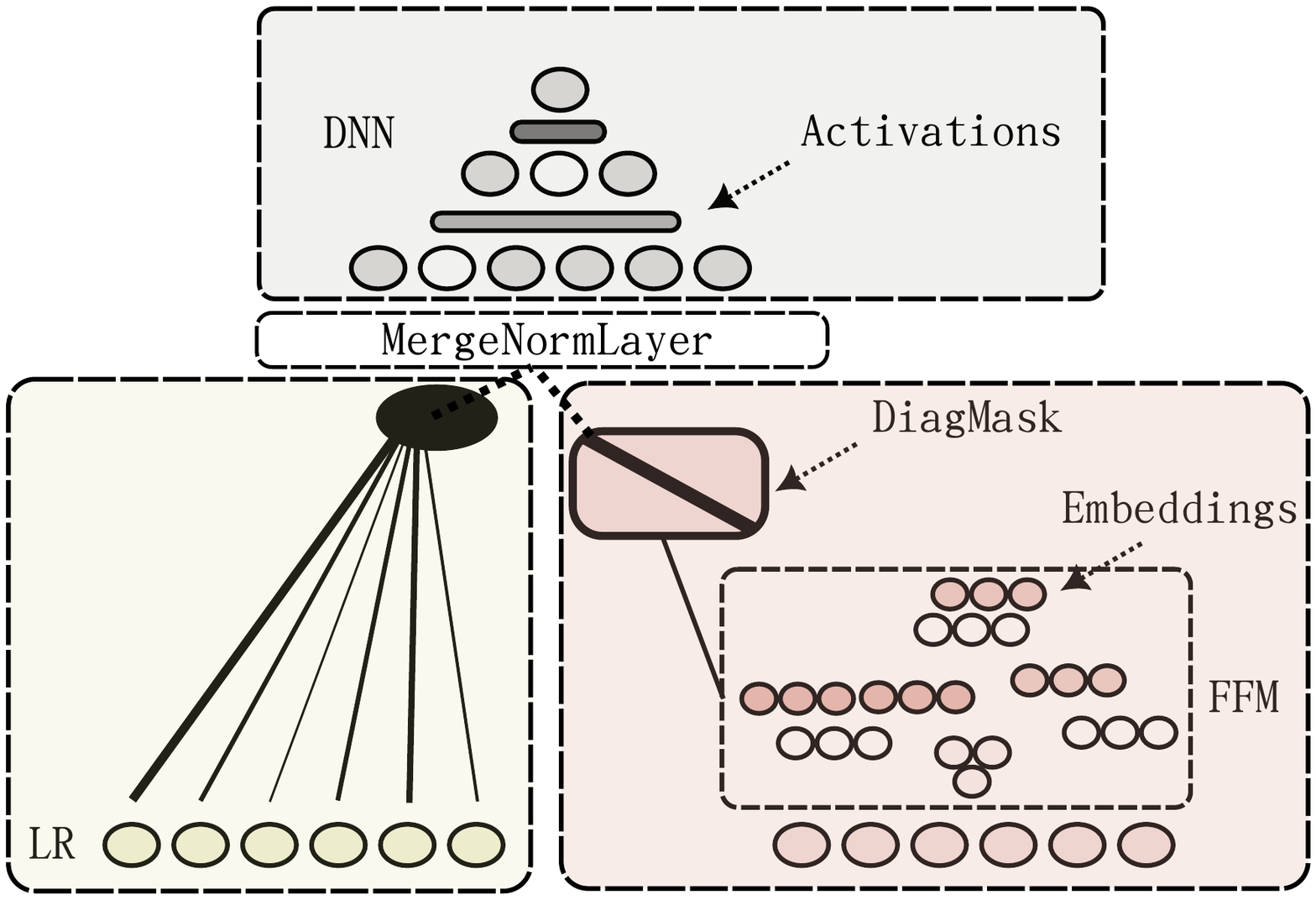}
    \caption{Architecture of implemented CPU-based DeepFFMs. Main blocks are the neural network (gray), logistic (yellow) and FFM (red) ones.}
    \label{fig:dffm}
    \vspace{-0.3cm}
\end{figure}

Despite these challenges, our deep learning-extended FFM method demonstrated significant performance gains over other tested algorithms in internal assessments. However, scaling this method was not straightforward. It was only through invoking BLAS~\cite{blackford2002updated}, that we achieved critical performance enhancements, allowing for practical full-scale deployment\footnote{\url{https://github.com/outbrain/fwumious_wabbit/blob/main/src/block_neural.rs}}. An overview of the architecture is shown in Figure~\ref{fig:dffm}.
. 
Key parts of the architecture are
\begin{align*}
    \textsc{lr}(w, x) = \sum_j^n w_j \cdot x_j + b; \textsc{ffm}(w, x) = \sum_{j_i = 1}^n \sum_{j_2 = j_1 + 1}^n (w_{j_1, f_2} &\cdot w_{j_2, f_1}) \\&\cdot x_{j_1} x_{j_2}.
\end{align*} Neural part (matrix form),
\begin{align*}
    \textsc{ffnn}(\textbf{W}_{1, 2, \dots, n}, \textbf{X}) = a_n (\dots a_2(a_1(\textbf{X} \cdot \textbf{W}_1) \cdot \textbf{W}_2) \dots )\cdot \textbf{W}_n,
\end{align*}
takes as input both FFM and LR's outputs, i.e. 
\begin{align*}
    \textsc{dffm}(\textbf{W}_{1, 2, \dots, n}, \textbf{w}_b, \textbf{w}_c, \textbf{x}) = &
    \textsc{ffnn}(\textbf{W}_{1, 2, \dots, n}, MergeNormLayer \\
    &(\textsc{lr}(\textbf{w}_b, x), 
    DiagMask(\textsc{ffm}(\textbf{w}_c, x))).
\end{align*}
Here, \emph{MergeNormLayer} represents the operator that combines outputs of FFM and LR parts and applies normalization. Further, \emph{DiagMask} represents diagonal mask of FFM space, inducing half smaller number of combinations requiring down-stream processing\footnote{See \url{https://github.com/outbrain/fwumious_wabbit/blob/main/src/regressor.rs} for more details.}.

\subsection{Criteo, Avazu and KDD2012 - a benchmark and stability analysis}
\begin{figure*}[htb!]
    \centering
    \includegraphics[width=.98\linewidth]{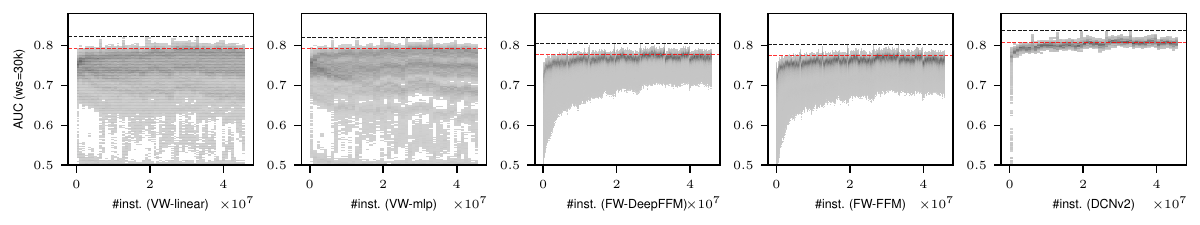}
    \includegraphics[width=.98\linewidth]{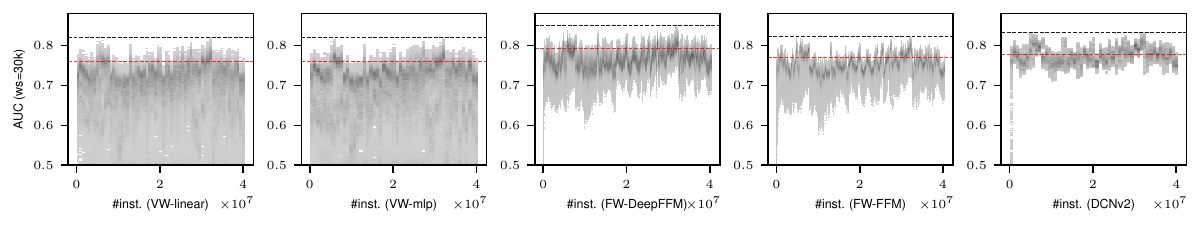}
    \includegraphics[width=.98\linewidth]{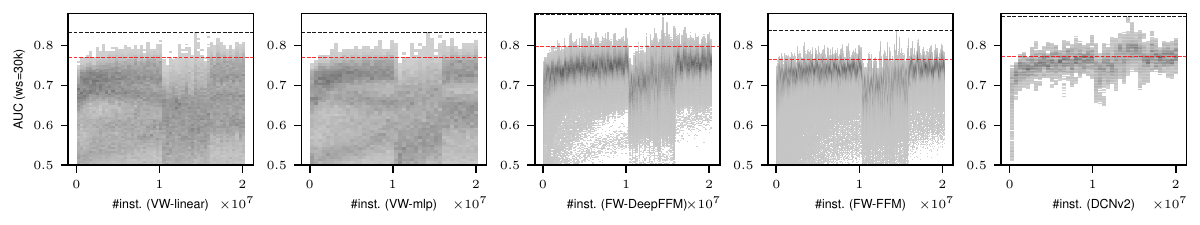}
\caption{Visualization of overall performance of different algorithms (single-pass) across different benchmark data sets (top-down: Criteo, Avazu, kddcup2012. Visualizations show traces of all trained models (per engine).}
\label{fig:auc-benchmark}
\end{figure*}

Even though we evaluated FW extensively on internal data sets (and online, in A/B tests), where it showed consistent dominance, results on published data sets such as Criteo are also of relevance for dissemination of engines' behavior and overall performance. In this section we overview a benchmark we conducted to assess general behavior of VW and FW. We also implemented DCNv2~\cite{wang2021dcn,shen2017deepctr}, a Tensorflow-based strong baseline\footnote{Unique hash was assigned to each value for this baseline for ease of implementation.}. For considered data sets (Criteo\footnote{\url{https://www.kaggle.com/c/criteo-display-ad-challenge}}, Avazu\footnote{\url{https://www.kaggle.com/c/avazu-ctr-prediction/data}} and KDD2012\footnote{\url{https://www.kaggle.com/c/kddcup2012-track2}}), log transform of continuous features was conducted and no additional data pruning (rare values etc.) was conducted (as is done in our system)\footnote{Such minimal pre-processing is within reach of a regular production.}. The hyperparameters considered include power of t, learning rates for different types of blocks (ffm, lr), regularization amount (L2 norm, VW). For DCNv2 we considered different learning rates, cross layer numbers, dropout rates and beta parameters.
Results of the benchmark are summarized in Figure~\ref{fig:auc-benchmark}. For each data set, algorithms considered are visualized as AUC scores computed in a rolling window of 30k instances\footnote{RIG and Log-loss scores are aligned with AUC-based results, hence only these are reported for readability purposes}.

The trace in each plot represents the average performance (95\% CI), and light-gray regions represent model evaluations that were out-of-distribution -- this aspect is particularly relevant for understanding \textbf{stability} of different approaches and their sensitivity to hyperparameter configurations. For example, we observed that adding deep layers to VW models in most cases resulted in worse performance. Carefully tuned VW hyperparameters yielded sufficient performance, however, indicate potentially cumbersome model search (when considering new use cases/data) in practice. Similar behavior was observed for DCNv2. The dotted black lines represent the overall best single-window performance, and performance on a given data set's test set\footnote{for KDD, we took last 2m instances to capture apparent variability in data better, other data sets are split as reported in their origin publications.}
Overall, initial phases of learning revealed VW's capability to adapt with less data, the DeepFFMs dominate after enough data is seen by the engines. Superior performance was observed by DCNv2 on Criteo, yet not other data sets (all features considered). The benchmark demonstrates that progressively more complex architectures tend to result in better modeling capabilities, and with them, better AUCs in this benchmark. 
In terms of \textbf{runtime}, on the same hardware, Criteo data set could be processed on average in 32min by VW, and 31min by FW (linear model vs. DeepFFM). Deep VW variations took substantially longer, around 65min on average (batch size of 2k). This result indicates that FW enables more powerful models with same time bounds for training. The DCNv2 (CPU) baseline was 30\%-50\% slower compared to DeepFFM runs. These tatistics were obtained based on tens of thousands of runs that represented different algorithm configurations (both hyperparameters and field specifications). Being CPU-based, the described approaches enable seamless scaling to commodity hardware, resulting in lower training and inference costs in practice.
\begin{table}
\let\center\empty
\let\endcenter\relax
\centering
\caption{Stability analysis and overall performance. Rows with max test set performance highlighted.}
Avazu (window=30k)
\resizebox{.9\width}{!}{\input{resultsAvazu}}
Criteo (window=30k)
\resizebox{.9\width}{!}{\input{resultsCriteo}}
KDDCup2012 (window=30k)
\resizebox{.9\width}{!}{\input{resultskddcup2012}}
\label{tab:bench}
\vspace{-0.6cm}
\end{table}
\section{FW in practice: Service Architecture overview}
\label{sec-architecture-overview}

This section aims to facilitate understanding of subsequently discussed optimizations that were put in place to enable scaling of Deep FFMs. The implemented FW contains both training and inference logic. The training logic is relevant for incrementally training \textbf{more than a hundred models}, online, every $n$ minutes (depends on the model). Training jobs are separate deployments that automatically query for relevant chunks of data, download, update based on existing weights and send the weights to the serving layer. Serving layer on-the-fly reconstructs the final inference weights via a patching mechanism discussed in Section 6, and exposes the weights as part of the serving service that handles millions of requests with new data.
Based on
the effect of predictions, data is streamed back to the system as training data (a feedback loop). The training jobs are Python-based services that interact with the binary via process invocations. Serving binds the inference capabilities with the serving (Java) service directly via a foreign function interface (ffi)\footnote{\url{https://github.com/outbrain/fwumious_wabbit/blob/main/src/lib.rs}}. The architecture enables separation of concerns -- training jobs are separate to inference jobs, albeit at the cost of needing to send the updated weight data between services; this is one of the key performance bottlenecks that was addressed in this work.

\section{Model training improvements}
\label{section-training-jobs}
We next discuss main improvements implemented at the level of training jobs and offline research.
\subsection{Speeding up model warm-up phase}
Model warm-up corresponds to a phase in model training where model starts with past data, and "catches up" with present data as fast as possible. We identified efficient data pre-fetching as a crucial optimization for speeding up this process. By implementing async learning cycles, multiple rounds of "future" data can be downloaded upfront, making sure the learning engine has constant influx of data.
Data pre-fetch in practice results in up to \textbf{4x faster pre-warming}. Within the cloud environment where the jobs are deployed, we can control machine "taints", i.e. signatures that determine their hardware profile. Pre-warm jobs have dedicated taints, which in practice results in machines that are newer and stronger.

\subsection{Hogwild-based training}
\label{sec-hogwild}
An optimization that significantly improved model pre-warm time is the previously reported Hogwild-based model training\cite{recht2011hogwild}, implemented also for Fwumious framework (as part of this work). Here, weight overlaps/overrides are allowed as the trade off for multi-threaded updates. By tuning Hogwild capacity to tainted machines, we observed multi-fold speedups in model warm-up. In practice, the times for bigger models went from multiple weeks to days, and in most cases around a day of training (to catch up). Weight degradation due to Hogwild was A/B tested and does not appear to cause any noticeable RPM drops. Summary of Howgild-based training compared to control (no such training) is shown in Table~\ref{tbl:hogwild}. Utilization of hogwild has shown substantial benefits also when utilized during online training (e.g., every 5min), and enabled scaling 100\% bigger models. To the best of our knowledge, this is one of the first demonstrations of consistent Hogwild-based training improvements for Deep FFMs.
\begin{table}[htb!]
    \centering
    \vspace{-0.2cm}
        \caption{Impact of Hogwild-based training.}
        \vspace{-0.3cm}
    \begin{tabular}{c|c}
      Implementation   & Warmup time (same period) \\ \hline
      FW-deepFFM-control   & 8d \\
      \rowcolor{Gray}
      FW-deepFFM-hogwild & 23h (48 threads) \\ \hline
      Implementation   & Online training (same period) \\ \hline
      FW-deepFFM-control   & 20m \\
      \rowcolor{Gray}
      FW-deepFFM-hogwild & 4m (4 threads)
    \end{tabular}
    \label{tbl:hogwild}
\end{table}
\subsection{Sparse weight updates}
\label{sec-sparse-updates}
The next discussed optimization is related to how gradients are accounted for during model optimization itself. We observed that deep layers, albeit being parameter-wise in minority compared to FFM part, take up considerable amount of time during optimization. To remedy this shortcoming, we identified an optimization opportunity that is a combination of activation function used in most models,
$f(x) = \textrm{max}(x, 0),$
and the specific implementation of FW. By realizing that we can identify \emph{zero global gradient} scenarios upfront, prior to updating any weights, we could skip whole branches of computation with no impact on learning. 
\begin{table}[b]
    \centering
    \caption{Speedups observed due to sparse weight updates.}
    \vspace{-0.2cm}
     \begin{tabular}{c|c|c|c|c}
        \hline \#Hidden layers & 1  & 2 & 3 & 4 \\ \hline \rowcolor{Gray}
        Speedup (sparse updates) &  1.3x  & 1.8x & 2.4x & 3.5x \\  \hline
    \end{tabular}
    \label{tab:hidden}
\end{table}
The performance (speed) of training, however, was across-the-board improved by 30\% for most models, and for deeper ones by \textbf{up to 3x}, see Table~\ref{tab:hidden} for more details. We observed that at most two hidden layers were feasible for production, hence any further speedups than observed 30\% were not feasible in practice. This optimization was possible due to ReLU's nature; this activation maps weights to zeros, effectively enabling identification of compute branches that need to be skipped during updates.

\section{Model serving improvements}
\label{sec:inference-part}
\begin{figure}[t]
    \centering
    \fbox{\includegraphics[width=\linewidth]{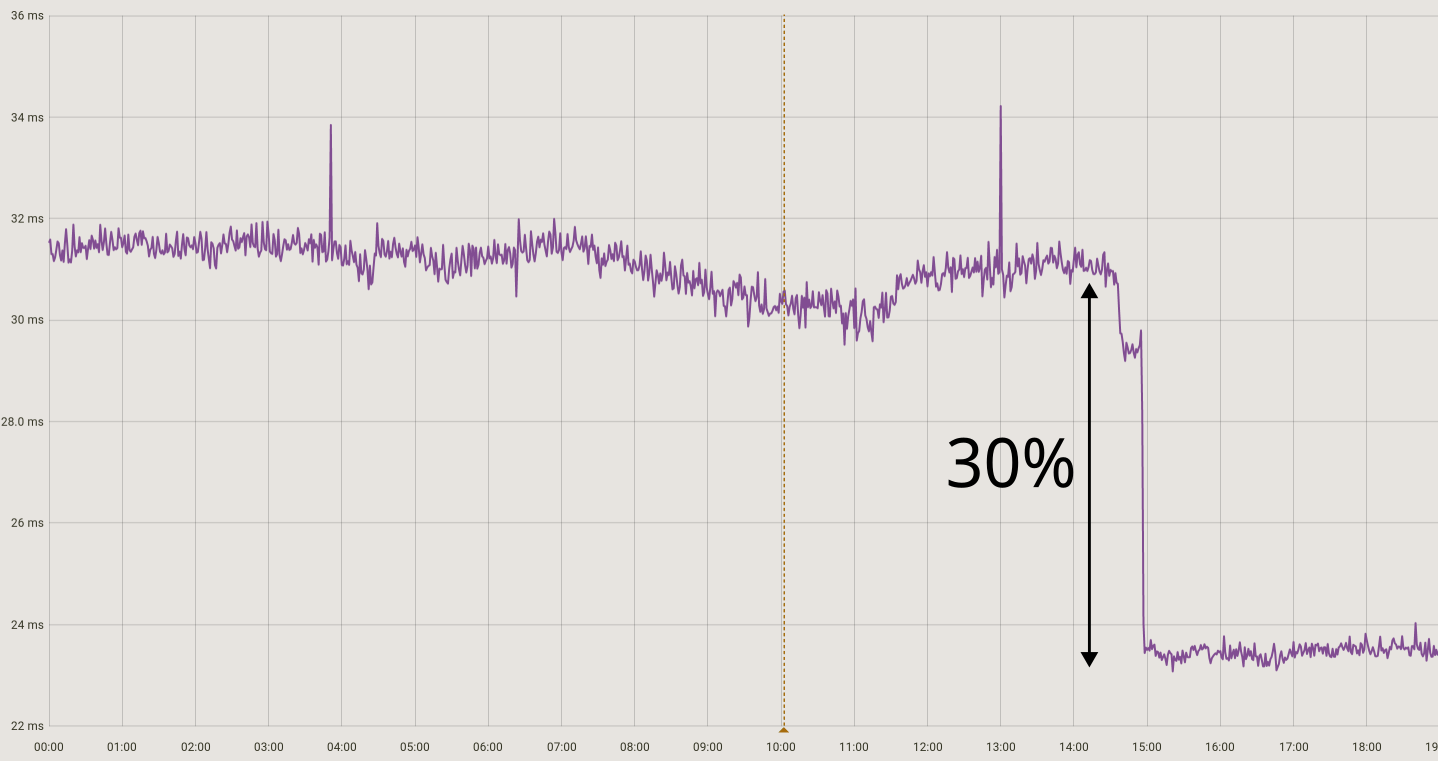}}
    \caption{Impact of context caching on inference time.}
    \label{fig:context-caching}
    \vspace{-0.4cm}
\end{figure}

\begin{figure}[!t]
    \centering
    \fbox{\includegraphics[width=\linewidth]{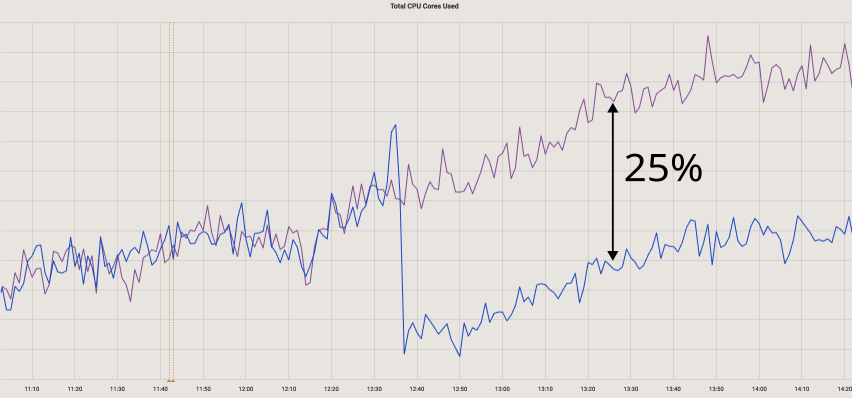}}
    \caption{Relative impact of SIMD-enabled (blue, after drop) vs. SIMD-disabled (purple) FW in production (inference).}
    \label{fig:simd}
    \vspace{-0.4cm}
\end{figure}

A considerable optimization we observed could take place in our system is \emph{context caching}. Each request can be separated into context and candidates. For all candidates in the request, the context is the same, even though the recommended content's features differ -- this implies part of the feature space is very consistent for each candidate batch. To exploit this property, a dedicated serving-level caching scheme was put in place. FW at this point does an additional pass only with the context part, where it identifies and caches frequent parts of the context. On subsequent candidate passes it reuses this information on-the fly instead of re-calculating it for each context-candidate pair. Deployment impact of context caching is shown in Figure~\ref{fig:context-caching}\footnote{\url{https://github.com/outbrain/fwumious_wabbit/blob/main/src/radix_tree.rs}}.
We next discuss \textbf{(SIMD) Instruction-aware forward pass}.
\label{sec:inference-simd}
Another optimization that is particular to inference is proper exploitation of SIMD intrinsics. These hardware instruction level optimizations, however, needed to be carefully implemented as the space of serving hardware is not homogeneous, meaning that on-the-fly instruction detection, and subsequent utilization of appropriate binary needed to be put in place. \textbf{SIMD intrinsics} were successfully used to speed up forward pass (inference) with no loss in RPM performance, and resulted in a consistent 20\% speedup for all serving\footnote{\url{https://github.com/outbrain/fwumious_wabbit/blob/main/src/block_ffm.rs}}. Real-life example of deployed SIMD-based FW vs. the control (no SIMD) is shown in Figure~\ref{fig:simd}. Up to 25\% faster inference (and with it lower resource utilization) were observed.

\section{Storage and transfer optimization}
As discussed in previous sections, training and serving jobs are separated. This separation of concerns, albeit easier to maintain, contributes to a major drawback: weight sending across the network. Model weights need to be constantly updated, which incurs substantial bandwidth costs. For example, \textbf{hundreds of live models} that take up to 10G of memory (per update) are constantly transferred across the network, resulting in a substantial bandwidth overhead to ensure low-latency online serving.

\textbf{Model patching}.
The first improvement we implemented is the concept of \emph{model patching}. This process is inspired by application of software patches (in general), albeit tailored to internal structure of FW's weights. Each trained model consists of training weights and the optimizer's weights. The latter are not required for actual inference, which immediately reduces the required space by half. Further, each subsequent inference weights update (inference weights can be multiple GB) first computes \emph{model diff} -- byte-level difference between old and new weights. This is possible due to a consistent memory-level structure of weight files. The diffs are compressed, sent to the serving layer, unpacked and applied to previous weights file to obtain the new set of weights (inference). This process takes tens of seconds, however, further reduces memory footprint on the network by more than 100\% (less than a GB of updates per model after patching Deep FFMs).

First, instead of storing absolute indices of bytes that change, \emph{relative} locations are stored, resulting in a considerable storage saving. Next, small integers denoting these differences are stored as a custom integer type -- instead of storing whole ints, compressed versions (small ints are impacted the most) are stored, leading to further improvements\footnote{\url{https://github.com/outbrain/fwumious_wabbit/blob/main/weight_patcher}}.
%Shematic overview of patcher is shown in Figure~\ref{fig:patcher}.
%\begin{figure}[t!]
%    \centering
%    \includegraphics[width=.7\linewidth]{patcher.pdf}
%    \caption{Schematic overview of model patcher algorithm. It identifies byte-level differences and stores them in a compact data structure that is sent instead of full weights.}
%    \label{fig:patcher}
%    \vspace{-0.5cm}
%\end{figure}
As patcher works at the level of bytes, we also successfully tested it for internal Tensorflow-based flows (reduced bandwidth for sending models).
Inspired by recent weight quantization advancements in the field of large language models~\cite{rokh2022comprehensive,bai2022towards}, we implemented a \textbf{variation of 16b weight quantization} that, when combined with the byte-level patching mechanism, offered considerable bandwidth and model storage improvements. The quantization algorithm was designed to account for the following use-case specific properties. First, by ensuring consistently small weight patches, the quantization ensures consistently smaller network load. Second, the quantization and \textbf{dequantization} procedures must be fast, as they need to happen within a designated time window after each training round (procedure has tens of seconds at most at its disposal for full weight space). Finally, the algorithm needs to be able to dynamically select viable weight ranges, as we observed considerable variation in weight update sizes based on e.g., time of the day (traffic amount). 
The final version of the algorithm can be summarized as follows. For each online model update (e.g., \textbf{5min window}), weights are first traversed to obtain the minimum and maximum values (weights). These statistics are required to dynamically determine the range of relevant weight bins, as the amount of possible values for 16b representation is small (around 65k). 
Let $W = \{w_1, w_2, \dots, w_n | w_i \in \mathbb{R}\}$ denote the set of all ($n$) weights and $b_{\textrm{max}}$ denote the number of possible weight buckets.
Once the minimum and maximum are obtained, the bucket size is computed as
$$
    \textsc{bucket}_s = \frac{\textrm{max}(W).\textrm{round}(\alpha)- \textrm{min}(W).\textrm{round}(\beta)}{b_{\textrm{max}}}.
$$
Note that minimum and maximum are \emph{rounded} to $\alpha$ and $\beta$ decimals. This consideration stems from empirical results that indicated that considering full precision bounds results in less stable patch sizes~\footnote{(quantization output tended to fluctuate more)}. When constraining minimum and maximum to certain precision, behavior stabilized whilst preserving performance and online behavior.
In the second pass, weights are \textbf{quantized} -- for each weight, its 16b representation is computed and stored. This results in computing
\begin{align*}
    &((w_i - \textrm{min}(W) / \textsc{bucket}_s)
    .\textrm{round}()
    .\textrm{castTo16b}()
    .\textrm{convertToBytes}(),
\end{align*}
i.e. a set of bytes that represent a certain weight bucket. Bytes are stored in FW weight format and re-used during inference. An important detail also concerns \emph{metadata} required to perform this type of quantization; the original weights file is enriched with a header that contains the bucket size and weight minimum -- these two properties are sufficient for efficient weight reconstruction when/where relevant\footnote{\url{https://github.com/outbrain/fwumious_wabbit/blob/main/src/quantization.rs}}.
Results on a representative CTR model are shown in Table~\ref{tab:quantization-results}. Metrics of interest are time to produce patch and the final patch/weight update's size. Patching and quantization result in up to \textbf{30x smaller model updates}.
\begin{figure}[htb!]
    \centering
    \fbox{\includegraphics[width=\linewidth]{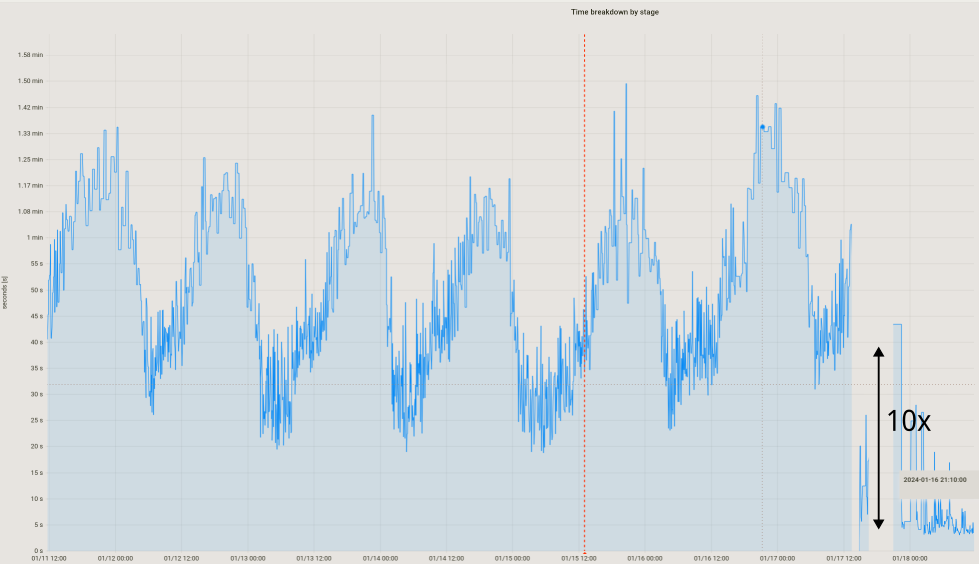}}
    \caption{Speedup observed when jointly using quantization and model patching (as opposed to just patching).}
    \label{fig:combined}
\end{figure}
\begin{table}[t]
    \centering
    \caption{Impact of model quantization on the global production CTR model.}
    \vspace{-0.2cm}
 \resizebox{.5\textwidth}{!}{%
    \begin{tabular}{c|c|c}
      Weight processing & Avg. time spent   & Update file size  \\ \hline
      no procecssing (baseline) & /  &  100\% \\
      fw-quantization & 2s & 50\% \\
      fw-patcher & 45s & 30$\pm 5$\% \\
      \rowcolor{Gray}
      \textbf{fw-patcher + fw-quantization} & \textbf{8s} & \textbf{3$\pm$2\% }
    \end{tabular}
}
    \vspace{-0.4cm}
    \label{tab:quantization-results}
\end{table}

Note that weight patching and quantization on their own already at least halve the size of weights that are used in serving and production. Further, by combining the two approaches, we observed a non-linear improvement in patch sizes -- around \textbf{10x smaller updates} are regularly produced. The quantized patches-based model showed small lifts in and online A/B against control with no quantization applied, considerably reducing network bandwidth required with a small positive business impact ($+ 0.15\%$ RPM). Speedup in a real-life production system due to compound effect of quantization and patching can be observed in Figure~\ref{fig:combined}. Rightmost part of the plot represents total time spent patching and computing quantized weights.

\section{Conclusions and open problems}
\label{sec:sec-open}
In this paper, we presented a collection of implementation details for scaling CPU-based DeepFFMs to operate at a multi-data-center scale, capable of handling hundreds of millions of predictions per second. We delved into both the offline and online components of our system. In the offline phase, we covered the complete workflow, including model architecture, enhancements to system warm-up processes, and bandwidth optimization strategies. Within the online phase, we describe two novel modifications to the inference layer that have yielded significant speed improvements. Our main algorithms, concepts, and performance benchmarks were discussed in detail, open-source implementations of key components were made freely available. The implementation is extensible to other FFM-based variants.
As further work, on the inference side, implementing quantization techniques could accelerate the forward pass by using integer-based operations~\cite{jacob2018quantization}. Improved weight sharing and memory mapping could offer training improvements.

\bibliography{sample-base}
\bibliographystyle{abbrv}
\end{document}

%% file: resultsAvazu.tex
\begin{tabular}{lrrrrr|r}
\toprule
algo & avg & median & max & std & min & test \\
\midrule
VW-linear & 0.6832 & 0.7016 & 0.8200 & 0.0668 & 0.4664 & 0.7596 \\
VW-mlp & 0.6755 & 0.6984 & 0.8200 & 0.0748 & 0.4664 & 0.7596 \\
\rowcolor{Gray} FW-DeepFFM & 0.7648 & 0.7654 & 0.8507 & 0.0243 & 0.4764 & 0.7916 \\
FW-FFM & 0.7524 & 0.7524 & 0.8234 & 0.0227 & 0.4816 & 0.7693 \\
DCNv2 & 0.7750 & 0.7745 & 0.8326 & 0.0202 & 0.5005 & 0.7763 \\
\bottomrule
\end{tabular}

%% file: resultsCriteo.tex
\begin{tabular}{lrrrrr|r}
\toprule
algo & avg & median & max & std & min & test \\
\midrule
VW-linear & 0.7340 & 0.7460 & 0.8219 & 0.0556 & 0.4768 & 0.7920 \\
VW-mlp & 0.7247 & 0.7425 & 0.8211 & 0.0670 & 0.4768 & 0.7920 \\
FW-DeepFFM & 0.7655 & 0.7689 & 0.8053 & 0.0179 & 0.4796 & 0.7803 \\
FW-FFM & 0.7578 & 0.7621 & 0.8020 & 0.0198 & 0.4682 & 0.7742 \\
\rowcolor{Gray} DCNv2 & 0.8042 & 0.8052 & 0.8370 & 0.0118 & 0.4958 & 0.8085 \\
\bottomrule
\end{tabular}

%% file: resultskddcup2012.tex
\begin{tabular}{lrrrrr|r}
\toprule
algo & avg & median & max & std & min & test \\
\midrule
VW-linear & 0.6333 & 0.6419 & 0.8336 & 0.0807 & 0.3430 & 0.7688 \\
VW-mlp & 0.6309 & 0.6402 & 0.8336 & 0.0869 & 0.3759 & 0.7688 \\
\rowcolor{Gray} FW-DeepFFM & 0.7323 & 0.7400 & 0.8781 & 0.0414 & 0.3687 & 0.7967 \\
FW-FFM & 0.7228 & 0.7318 & 0.8382 & 0.0391 & 0.3651 & 0.7641 \\
DCNv2 & 0.7589 & 0.7610 & 0.8718 & 0.0301 & 0.4792 & 0.7734 \\
\bottomrule
\end{tabular}